\documentclass[letterpaper, 10 pt, conference]{ieeeconf}  % Comment this line out if you need a4paper

\IEEEoverridecommandlockouts                              % This command is only needed if 
\usepackage{graphics} % for pdf, bitmapped graphics files
\usepackage{epsfig} % for postscript graphics files
\usepackage{mathptmx} % assumes new font selection scheme installed
\usepackage{times} % assumes new font selection scheme installed
\usepackage{amsmath} % assumes amsmath package installed
\usepackage{amssymb}  % assumes amsmath package installed
\usepackage{bm}
\usepackage{algorithm}
\usepackage{multirow}
\usepackage{verbatim}
\usepackage{graphicx}
\usepackage{hyperref}

\usepackage{subcaption}
\usepackage{lipsum}
\usepackage{mwe}
\usepackage{booktabs} % For formal tables.
\usepackage{cite}

\usepackage{titlesec}
\titlespacing{\section}{0pt}{*0.4}{*0.4}
\titlespacing{\subsection}{5pt}{*0.2}{*0.2}
\usepackage[font=footnotesize]{caption}
\title{\LARGE \bf
Metrics for Evaluating Social Conformity of Crowd Navigation Algorithms$^{*}$
\thanks{$^{*}$This project is supported by the Foundation for Australia-Japan Studies under the Rio Tinto Australia-Japan Collaboration Project.}
}

\author{Junxian Wang$^{1}$, Wesley P. Chan$^{1,2}$, Pamela Carreno-Medrano$^{1}$, Akansel Cosgun$^{1}$, Elizabeth Croft$^{1}$
\thanks{$^{1}$Monash University}
\thanks{$^{2}$Corresponding author}
} %\author
\begin{document}

\maketitle
\thispagestyle{empty}
\pagestyle{empty}

%%%%%%%%%%%%%%%%%%%%%%%%%%%%%%%%%%%%%%%%%%%%%%%%%%%%%%%%%%%%%%%%%%%%%%%%%%%%%%%%
\begin{abstract}
Recent protocols and metrics for training and evaluating autonomous robot navigation through crowds are inconsistent due to diversified definitions of ``social behavior". This makes it difficult, if not impossible, to effectively compare published navigation algorithms. Furthermore, with the lack of a good evaluation protocol, resulting algorithms may fail to generalize, due to lack of diversity in training. To address these gaps, this paper facilitates a more comprehensive evaluation and objective comparison of crowd navigation algorithms by proposing a consistent set of metrics that accounts for both efficiency and social conformity, and a systematic protocol comprising multiple crowd navigation scenarios of varying complexity for evaluation. We tested four state-of-the-art algorithms under this protocol. Results revealed that some state-of-the-art algorithms have much challenge in generalizing, and using our protocol for training, we were able to improve the algorithm's performance. We demonstrate that the set of proposed metrics provides more insight and effectively differentiates the performance of these algorithms with respect to efficiency and social conformity.

\end{abstract}

%%%%%%%%%%%%%%%%%%%%%%%%%%%%%%%%%%%%%%%%%%%%%%%%%%%%%%%%%%%%%%%%%%%%%%%%%%%%%%%%
\section{INTRODUCTION}
        \label{sec:introduction}
        Robot navigation in crowds is a growing research area \cite{malaysurvey, rlsuperiority, humansurvey}. Design of algorithms allowing robots to navigate safely and socially in populated environments an open challenge. Robots must be able to navigate cooperatively with humans while reasoning about their own actions' impact on surrounding humans and have a good understanding of the ongoing interactions between these humans. Furthermore, robots must follow a range of social and cultural rules \cite{humansurvey}, such as respecting personal spaces, and keeping to a specific side preference when passing other humans. As social norms are often contextual, they are difficult to define such that they can be explicitly implemented into algorithmic rules.

Recently, advancements in computation hardware and machine learning have enabled a series of crowd navigation algorithms based on deep reinforcement learning \cite{towards,sacadrl,cadrlv2,sarl,rgl,mapless,dwarl,structuralrnn,densecavoid,crowdsteer}. In these approaches, social behavior is encoded in the reward function, which provides a penalty or reward based on the robot's behavior. However, a survey of the latest papers on crowd navigation algorithms reveals diversified definitions and metrics for what constitutes ``social behavior". Although all papers consistently report on task-related metrics such as success rate or time to goal, few include quantitative measures for social conformity of the robot's behavior. While efficiency is an important aspect for successful navigation in crowds, social conformity is equally important \cite{humansurvey}. Furthermore, when training and evaluating their algorithms, some of these works only included a handful of navigation scenarios (as few as one) \cite{cadrlv2,sarl,rgl}. This risks the resulting algorithm being unable to generalize to different scenarios, and its performance deteriorating greatly with slight changes in the environment.

Given the lack of consistency in social evaluation metrics used for crowd navigation algorithms, it is not surprising to find that some papers that aim to provide socially conforming crowd navigation algorithms present evaluations that are mainly focused on task efficiency, and fail to detail the social conformity of the learned navigation policy. With different papers evaluating algorithms using different navigation scenarios, assessing how well the algorithms translate to different navigation scenarios is challenging. The lack of well-defined standardized metrics and evaluation protocols inhibits the comparison of performance among published algorithms. This paper addresses this gap by proposing a set of metrics, focusing on social conformity, as well as an evaluation protocol, incorporating a range of common navigation scenarios, that can be used to objectively evaluate and compare different crowd navigation algorithms. The proposed metrics are grounded in evaluation criteria and aspects of social conformity found in existing literature. We then apply our proposed set of metrics on a collection of state-of-the-art crowd navigation algorithms. In the process, we found that the performance of these algorithms can deteriorate greatly when evaluated in scenarios that are slightly different from the one it was trained on. Hence, we also retrained these algorithms using our proposed protocol, and demonstrate improved performace across a range of scenarios.

The main contributions of this paper include: \textbf{i)} a set of specifically designed metrics that focus on evaluating both social conformity and efficiency of robot navigation behaviors, \textbf{ii)} a systematic and comprehensive evaluation protocol comprising multiple crowd navigation scenarios of varying complexity, \textbf{iii)} a demonstration of improved performances of state-of-the-art crowd navigation algorithm through application of our proposed protocol for training, and \textbf{iv)} application of our proposed metrics and evaluation protocol on, as well as in-depth comparison of, a collection of the most recent crowd navigation algorithms. 
        
\section{RELATED WORK}
        \label{sec:related_works}
        %A review of existing literature in crowd navigation \cite{towards,sacadrl,cadrlv2,sarl,rgl,mapless,dwarl,structuralrnn,densecavoid,crowdsteer} reveals that a diversified range of evaluation metrics is used to assess robot performance. Although all papers consistently report on task-related metrics such as success rate or time to goal, most include few or no quantitative measures of the social conformity of the robot's behaviors. While efficiency is one important aspect for successful navigation in crowded environments, social conformity is also equally important \cite{humansurvey}.

Existing works tend to focus on task-related metrics when evaluating crowd navigation algorithms. For instance, Liu et al. \cite{structuralrnn} and Everett et al. \cite{cadrlv2} use time to goal as their evaluation metric. In \cite{cadrlv2} only circle crossing scenarios were used to evaluate the algorithm, where as in \cite{structuralrnn}, an additional group scenario was used. Sathyamoorthy et al. \cite{densecavoid}, uses trajectory length rather than time to goal, a similar metric. They additionally have a freezing robot percentage as their criteria, which they argue offers a more fine-grained success rate metric. These papers rely on using efficiency oriented metric for evaluating their navigation methods, and do not include any metrics for evaluating social conformity. 

Among the recently emerging papers in which some notion of social conformity is considered during evaluation, Patel et al.'s paper presenting the DWA-RL (Dynamic Window Approach Reinforcement Learning) algorithm \cite{dwarl} and Liang et al.'s work presenting the CrowdSteer algorithm \cite{crowdsteer} propose the robot's average velocity and the smoothness of the robot's velocity as measures of the naturalness of the robot's trajectory. They later use these metrics to evaluate the robot's ability to navigate at a socially comfortable speed. 

In the papers presenting the OM-SARL (Socially Aware Reinforcement Learning with Occupancy/Local Mapping) algorithm \cite{sarl}, RGL (Relational Graph Learning) algorithm \cite{rgl}, and SA-CADRL (Socially Aware Collision Avoidance with Deep Reinforcement Learning) \cite{sacadrl} algorithms, proximity to other pedestrians was used as a component of their metric\footnote{OM-SARL and RGL include the average reward obtained as a metric. Since the reward function includes a proximity penalty, the metric indirectly accounts for the robot's proximity to pedestrians.}. This metric measures the robot's ability to respect nearby people's personal space during navigation. Chen et al.~\cite{sacadrl} also use metrics for specific social norms, including the percentage of times that the robot passes, overtakes, and crosses other pedestrians on the same preferred side (left or right). Their proposed algorithm was trained with the metrics related to these specific social norms built into the reward function. The papers \cite{sarl, rgl} only evaluated their algorithm in a circle crossing navigation scenario.

Jin et al.'s work \cite{mapless} not only includes the number of times a robot violates other people's personal space, but also has a metric for social zone violations. The authors define a social zone to be a projected rectangular area, with a length proportional to their speed, in front of the humans and the robot. %, with the length of the rectangle being proportional to their speed. 
This metric measures how safe the robot's motion appears to other humans by summing the duration of time when the robot's social zone intersects with other humans'.

% Our review shows that these navigation algorithms have different definitions of performance and evaluation metrics; hence, they cannot be readily compared with each other. This paper aims to address this issue by proposing a fundamental set of metrics that targets the social conformity of crowd navigation algorithms to allow easier and more in-depth comparison among current and future crowd navigation methods.

Recently, researchers have identified the need to improve the evaluation of crowd navigation algorithms. For instance, Tsoi et al. \cite{sean} introduce a simulation platform including multiple real-life scenarios. However, the simulated pedestrians are not able to react to the robot's actions. Thus, they do not allow quantification of the robot's actions' impact on surrounding humans. Biswas et al. \cite{navbench} proposed a set of metrics for evaluating crowd navigation algorithms; nonetheless, their metrics remain mostly focused on robot efficiency rather than social conformity.

\section{PROPOSED EVALUATION METRICS}
        \label{sec:method}
        We propose a set of metrics consisting of both basic performance metrics and social conformity-focused metrics. The basic performance metrics include success rate, collision percentage, time-out percentage and the robot's average navigation time. For the social conformity-focused metrics, Kruse et al.\cite{humansurvey} define three different aspects of social conformity that should be considered when evaluating a robot navigation behavior: \textbf{comfort}, \textbf{naturalness} and \textbf{sociability}. For each aspect, we first provide a description. We then define the proposed metric(s) belonging to that aspect, and provide a detailed explanation of the rationale behind each metric. Table ~\ref{tab:metrics} summarizes the proposed social conformity-focused metrics.

\begin{table}[t]
\centering
\vspace{1mm}
\caption{Summary of the proposed metrics}
\label{tab:metrics} 
\setlength\tabcolsep{2pt}
    \begin{tabular}{|c|p{1.3cm}|p{5.0cm}|c|}
    \hline
    No. & Name    & Description    & Category    \\ \hline                                                 
    \textbf{I}             
        & Personal Space  
        & Average duration of robot being inside the min. comfortable personal space
        & Comfort     \\ \hline
    \textbf{II}            
        & Projected Path  
        & Average duration of when robot's projected path intersects with a human's projected path  
        & Comfort     \\ \hline
    \textbf{III}           
        & Aggregated Time 
        & Aggregated goal reaching time for all cooperative agents
        & Comfort     \\ \hline
    \textbf{IV}            
        & Integrated Jerk 
        & Average squared jerk over the robot's trajectory (c.f. Min. Jerk Trajectory humans employ)
        & Naturalness \\ \hline
    \textbf{V}             
        & Walking Speed   
        & Average duration of when robot's speed exceeds human max. normal walking speed
        & Naturalness \\ \hline
    \textbf{VI}            
        & Side \newline Preference 
        & Side preference (left or right) percentage for passing/overtaking/crossing behavior          
        & Sociability \\ \hline
    \end{tabular}
\end{table}

\subsection{Notation}
In this paper, $r$ denotes the robot, $h_n$ denotes the $n-\text{th}$ human, and $H$ denotes the set of all humans. Superscript denotes either belonging to robot $r$ or human $h$, subscript $x$ and $y$ denotes x and y-coordinates, subscript $t$ denotes time $t$. Let $T$ be total amount of time taken in an episode. We denote with $\boldsymbol{p^a}=[p_{x}^a, p_{y}^a], \,  \boldsymbol{\vec{v^a}}=[v_{x}^a, v_{y}^a] \in \mathbb{R}^2$  the position (x,y coordinate) and velocity (x,y speed) of an agent (robot or human) $a$ respectively. Additionally, $M_i$ denotes the $i\text{th}$ metric, $rad$ denotes radius, $\bm{s}$ denotes starting position, and $\bm{g}$ denotes goal position.

% \begin{itemize}
%     \item \textbf{Comfort} is the absence of annoyance and stress for humans in interaction with robots
%     \item \textbf{Naturalness} is the similarity between robots and humans in low-level behavior patterns
%     \item \textbf{Sociability} is the adherence to explicit high-level cultural conventions
% \end{itemize}

\subsection{Comfort}

Kruse et al. \cite{humansurvey} define comfort as absence of annoyance and stress for humans when interacting with robots. Specifically, to increase comfort of surrounding humans, the robot must allow surrounding humans to have a perceived sense of safety from the robot. To provide this sense of safety, the robot should avoid entering pedestrians' personal space, avoid the apparent intended path of pedestrians, and plan a path that will impact pedestrians the least. Hence, we propose the following metrics for measuring comfort:
\begin{enumerate}
    \item[$M_I$] Average duration a robot spends inside the minimum comfortable personal space of pedestrians (radius $\epsilon$).
    \item[$M_{II}$] Average duration that a robot's immediate projected path intersects with the projected paths of pedestrians.
    \item[$M_{III}$] Aggregated goal reaching time for the robot and the pedestrians.
\end{enumerate}

The first metric, $M_I$, is on the robot's ability to respect personal spaces. Generally, the robot needs to trade-off between efficiency and respecting pedestrians' personal spaces. %This metric requires a clear definition of personal space. 
We define this metric as the total duration of when the distance between robot's and pedestrian's position is less than $\epsilon$, averaged over the total path time, with
$M_I$ calculated as:
\begin{equation}
    M_I = \frac{1}{T} \sum_{t=0}^{T} 
    \left \{ 
    \begin{matrix}
        1,       &\min_{h \in H}(|\bm{p^h_t} - \bm{p^r_t}|) < \epsilon \\
        0,       &\mathrm{otherwise}.
    \end{matrix}
    \right.
\end{equation}
Based on Hall's Proxemic model \cite{hidden} where a distance of 1.2-3.6m between strangers is prescribed, we suggest setting $\epsilon$=1.2 m for general cases. However, the size of comfortable personal space can depend on and be adjusted based on context such as location and culture. 

\begin{figure}
    \begin{center}
    \includegraphics[width=0.3\linewidth]{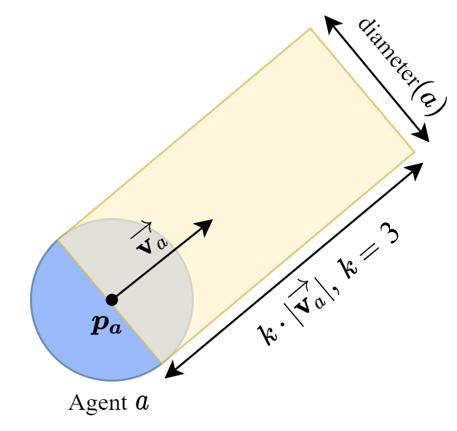}
    \caption{Social safety zone (also referred to as velocity rectangle) of agent $a$, defined for $M_{II}$}\label{fig:vel_rect}
    \end{center}
\end{figure}

The second metric, $M_{II}$, relates to the robot's motion intention. Humans reason about future trajectories of other pedestrians and mobile agents and make adjustments to avoid collisions. Thus, if a robot's trajectory appears to be intersecting with a human's path, and the time to collision is short enough such that it compels the human to change their own course, the human is likely to feel uncomfortable. To address this, \cite{mapless} suggested a social safety zone, where a rectangular section is projected in the direction of the agent's velocity, with length proportional to speed, and width the same as the agent's width. This area for an agent $a$, denoted by $vr^a$ (Fig. \ref{fig:vel_rect}), is called the velocity rectangle. We measure the discomfort caused by the robot's current heading as the aggregated duration during which the robot's velocity rectangle intersects with other humans', averaged over total path time. Formally, $M_{II}$ is defined as:
\begin{equation}
    M_{II} = \frac{1}{T} \sum_{t=0}^{T}
    \left \{ 
    \begin{matrix}
        1,       &\exists_{h \in H} \: \mathrm{intersect}(vr^r_t, vr^h_t) \\
        0,       &\mathrm{otherwise}.
    \end{matrix}
    \right.
\end{equation}

The last metric in this category, $M_{III}$, corresponds to the aggregated navigation time for all pedestrians and the robot. To allow nearby pedestrians to navigate comfortably the robot should exhibit cooperative path planning. Cooperative path planning aims to have all agents reach their goal with the shortest aggregated amount of time, rather than prioritizing a single agent's time to goal. Hence, $M_{III}$ measures the total amount of time spent for all agents in the environment to reach their goal:
\begin{equation}
    M_{III} = \sum_{a \in [r, H]} \mathrm{navTime}(a).
\end{equation}

\subsection{Naturalness}
Naturalness metrics relate to the similarity between robots and humans in low-level behavior patterns, such as speed. To evaluate the naturalness of robots, Kruse et. al.'s work \cite{humansurvey} considers the smoothness of the robot's trajectory. Furthermore, the robot should navigate at normal human walking speed when they operate in a crowd with other humans present. These two metrics are summarized as follows.
\begin{enumerate}
    \item[$M_{IV}$] Average squared jerk over the robot's trajectory
    \item[$M_{V}$] Average duration over the path that the robot's speed exceeds human maximum normal walking speed
\end{enumerate}
        
Based on the assumption that humans tend to follow a trajectory with near minimum jerk \cite{humansurvey}, i.e., a smooth trajectory, the fourth metric, $M_{IV}$, considers the similarity between the robot's and humans' trajectories in terms of smoothness. The integrated jerk over a trajectory is used to quantitatively measure robot path smoothness. $M_{IV}$ for a path $x(t)$ is calculated as:
\begin{equation}
    M_{IV} = \frac{1}{T} \sum_{t=0}^{T} \dddot{x}(t)^2, 
\end{equation}
where
\begin{equation}
    \dddot{x}(t)=\frac{d^3x(t)}{dt^3} 
\end{equation}
is computed in discrete time.

The fifth metric, $M_{V}$, accounts for the robot's navigation speed. Specifically, we measure the average time when the robot's action exceeds the empirically observed maximum walking speed of humans (i.e., 1.2-1.5 m/s \cite{walkingspeed}). %, to allow humans to make comfortable decisions on avoiding the robot if necessary. 
We calculate $M_{V}$ as:
\begin{equation*}
    M_{V} = \frac{1}{T} \sum_{t=0}^{T} 
    \left \{ 
    \begin{matrix}
        1,       &|\bm{\vec{v}_t}| > 1.5 \\
        0,       &\mathrm{otherwise}.
    \end{matrix}
    \right.
\end{equation*}

\subsection{Sociability}
This metric relates to an agent's ability to follow social rules or norms specific to particular societies or cultures. This could include passing, crossing and overtaking others on a specific side, and not cutting through a group of people. %Sociability can be analyzed by the robot's ability to follow a range of social norms. 
One such norm is originally from the pedestrian behavior model by Helbing \cite{mathmodel}, and adopted by SA-CADRL\cite{sacadrl}. This model suggests the robot to adhere to a specific side when passing, overtaking and crossing other people's path. Thus, we measure sociability as the side preference exhibited by a given path planning method when tested on passing, overtaking and crossing scenarios (see Fig. ~\ref{fig:scenarios} (5)-(7)): 

\begin{enumerate}
    \item[$M_{VI}$] Side preference (either left or right) for passing/overtaking/crossing behavior
\end{enumerate}

Based on the description of side preference given in \cite{sacadrl}, to compute this metric, in each scenario, the robot assumes a fixed start and goal position. As the robot approaches the person such that $p_{y}^r < p_{y}^h + rad^h + rad_r$, the case is labeled as ``left" if $p_{x}^h > p_{x}^r$ and ``right" if $p_{x}^h < p_{x}^r$. The percentage of ``left" and ``right" cases are then recorded.

%To compute this metric the following procedure was employed. In each scenario, the robot assumes a fixed start and goal position. As the robot approaches the person such that $p_{y}^r < p_{y}^h + rad^h + rad_r$, the case is labeled as ``left" if $p_{x}^h > p_{x}^r$ and ``right" if $p_{x}^h < p_{x}^r$. %The start position of the human is varied laterally between $\pm (\mathrm{rad}^r + \mathrm{rad}^h)$ in 20 equal increments, and the percentage of ``left" and ``right" cases are recorded. This is based on the description of side preference given in \cite{sacadrl}.

\subsection{Other considerations}
Another social norm that should be followed is to avoid passing through a group of people\cite{humansurvey}. This requires the robot to have the ability to identify groups. However, this aspect is, to certain extent, covered by the personal space metric, as a group of people usually stays within the personal distance to each other, leaving no gap for the robot to navigate through the group without invading other people's personal space (accounted for by $M_{I}$). Other additional social norms during navigation may include crossing streets only at pedestrian crossings, or letting people in an elevator come out first before entering. Such aspects are highly context-dependent, and hence are not included in the set of proposed fundamental metrics in this paper. 
        
\section{EVALUATION PROTOCOL}
        \label{sec:protocol}
        % Aside from the lack of a common set of evaluation metrics, our review of literature also revealed the lack of a common set of evaluation protocols, that is, different papers often evaluate their algorithms using different navigation scenarios. Hence, we also propose a set of common evaluation scenarios.  

\begin{figure*}
\centering
\vspace{1mm}
\includegraphics[width=.9\textwidth]{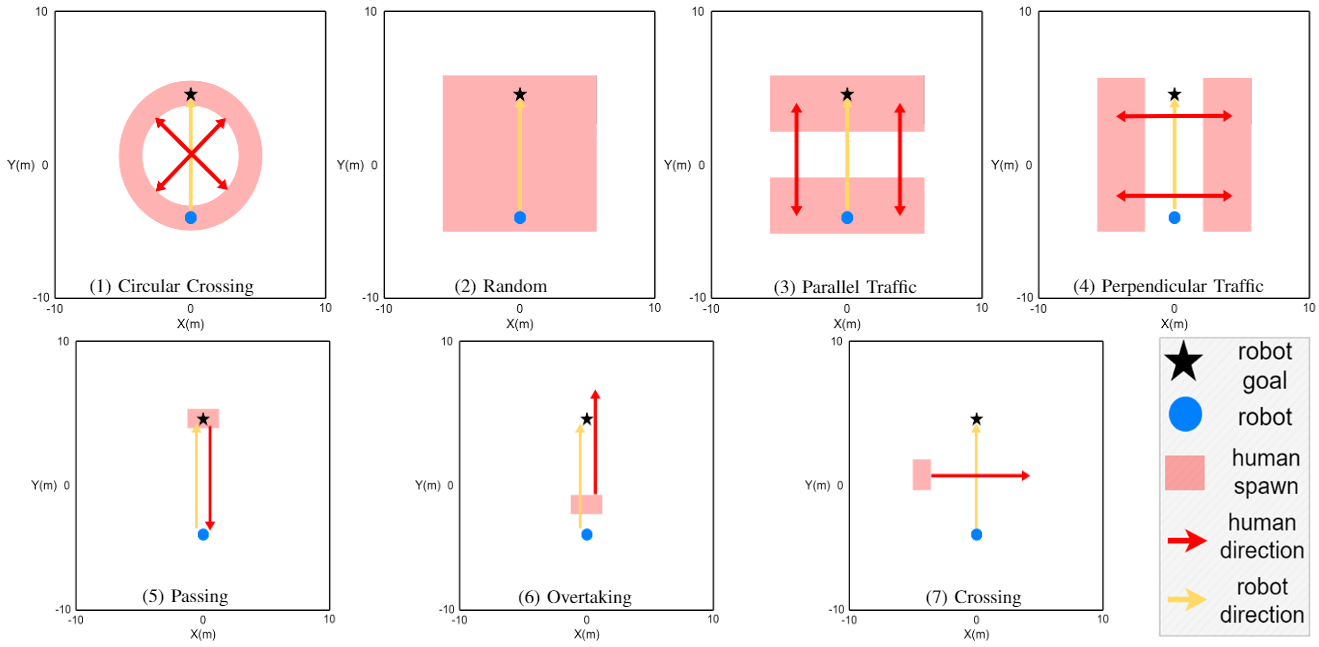}
\caption{The set of proposed scenarios. Red areas represent human spawn areas, Red arrows represent humans' travel direction, Yellow arrows represent robot's travel direction, Blue dot represents robot's starting point, Black star represents robot's destination}
\label{fig:scenarios}
\vspace{-4mm}
\end{figure*}

We propose a set of seven common scenarios for evaluating crowd navigation algorithms. The first four are designed to expose the agent to different multi-pedestrian navigation situations. The last three evaluate the consistency of the robot's side preference and thus include a single pedestrian. Our set of scenarios are listed below and illustrated by Fig. \ref{fig:scenarios}, with all having fixed start and target positions for the robot. The origin of the map is at the center. For the robot, $\bm{s^r} = \{0,\: s^r_y\}$, $\bm{g^r} = \{0,\: g^r_y\}$, $s^r_y = -g^r_y$.

\begin{enumerate}
    \item \textbf{Plaza - Circular Crossing}: Humans start at a random location on a ring and try to reach the other side of the ring. The ring has the same diameter as the distance between robot's start and goal. This is a scenario commonly used in existing works \cite{sarl, rgl, cadrlv2, structuralrnn}.
    \item \textbf{Random}: Pedestrians start at random locations and attempt to reach random target locations within a predefined square area of width 50\% larger than the distance between robot's start and goal locations. The robot travels through this randomly moving crowd. %The pedestrian area is a square and has width 50 percent larger than the distance between robot's start and goal.    
    \item \textbf{Parallel Traffic}: A pedestrian crowd moves parallel to the robot. There are two areas of possible pedestrian start locations, at the robot's start and goal side. Pedestrians start randomly in one area and their goals are placed randomly in the opposite area.
    \item \textbf{Perpendicular Traffic}: A pedestrian crowd moves perpendicular to the robot's direction of motion, similar to the parallel traffic scenario but the start and goal areas are on the left and right side of the robot's path.
    \item \textbf{Passing}: A pedestrian starts opposite the robot, and they move towards and pass each other. Specifically, the start and goal positions for human and robot are within $s^h_x, g^h_x \in [s^r_x - \mathrm{rad}^r - \mathrm{rad}^h, \quad s^r_x + \mathrm{rad}^r + \mathrm{rad}^h]$ and $s^h_y = g^r_y$, $g^h_y = s^r_y$.
    \item \textbf{Overtaking}: The robot starts behind a pedestrian (with slower speed) travelling in the same direction and overtakes the pedestrian. Specifically, the start and goal positions $s^h_x$ and $g^h_x$ are within the same range as the passing scenario, and $s^h_y > s^r_y$, $g^h_y > g^r_y$.
    \item \textbf{Crossing/Intersection}: One human moves perpendicularly to the robot's direction, and they cross paths with each other. Specifically 
    $s^h_y, g^h_y \in [- \mathrm{rad}^r - \mathrm{rad}^h, \quad \mathrm{rad}^r + \mathrm{rad}^h]$ and $s^h_x = -g^h_x = s^r_y$.
\end{enumerate}

% When evaluating crowd navigation algorithms in simulation, there are two settings of the environment: the robot is either visible or invisible to humans (both settings simulate humans using the ORCA (Optimal Reciprocal Collision Avoidance) \cite{orca} model). All metrics, except $M_{III}$ ``Aggregated Time", are tested in the invisible-robot environment (as most existing works do). This is because: \textit{(i)} $M_{III}$ only makes sense if the humans can see the robot, otherwise all algorithms will have the same result. \textit{(ii)} All metrics except $M_{III}$ are affected by the humans' motion and cannot fully evaluate the targeted model's performance when humans are also actively avoiding the robot. During evaluation, the results are taken from successful episodes out of a total of 500 episodes for each of the first four scenarios. For the single-agent scenarios, since they are used for evaluating side preference, which is of interest only when the human and robot get sufficiently close to each other at some point of their trajectories, the start configuration (i.e., the human start position) only needed to be varied within a small range as shown in Fig. \ref{fig:scenarios} (5)-(7), and side preference can be effectively evaluated in a much smaller number of test cases. In each of these scenarios, the human start position is varied among 20 equally spaced positions (in the direction perpendicular to robot travel) and evaluation is performed on these 20 cases.

\section{EXPERIMENT SETUP}
        \label{sec:experiment}
        \subsection{Tested Algorithms}
To demonstrate the effectiveness of the proposed metrics, we used them to evaluate four state-of-the-art crowd navigation algorithms that have made their code publicly available. We provide a brief description of each of these algorithms below. Interested readers are referred to the given references for details of each algorithm. 

\textbf{CADRL} \cite{cadrl}: One of the earliest algorithms that uses deep reinforcement learning for crowd navigation. It only considers one other human during planning. Multi-agent navigation is achieved by computing the output value from the model against each human, and selecting the action with the highest cumulative reward.

\textbf{GA3C-CADRL} \cite{cadrlv2}: This algorithm incorporates an LSTM (Long Short Term Memory) module to encode an arbitrary number of humans' state along with the robot's state as input to their A3C (Asynchronous Advantage Actor-Critic) framework. 
%The parameters specific to this model are taken directly from the original paper.

\textbf{OM-SARL} \cite{sarl}: This algorithm models robot-human interaction through the use of local maps. The aggregation of all maps, through a pooling module, is provided as the input to an RL framework. 

\textbf{RGL} \cite{rgl}: This algorithm learns the weights of a Relational Graph Neural Network by passing in a similarity matrix and a latent matrix containing the states of the robot and all humans. The relational graph can be used to infer the human trajectories taking into account the interactions between robot and humans, and among humans.

We first began evaluating the selected algorithms as originally published, using the opensourced code provided in each paper. However, we found that some of the algorithms perform much poorer than originally reported in the respective papers, even for standard metrics such as success rate. We suspected that this was due to the published algorithms having trained only in a limited number of scenarios. Hence, we decided to retrain all selected algorithms using our proposed protocol, and perform another evaluation.

\subsection{Training Environment}
We trained each algorithm in identical conditions, i.e., same training parameters and scenarios, using the first four scenarios in our proposed evaluation protocol. We used the CrowdSim simulator environment \cite{rgl}, and ORCA (Optimal Reciprocal Collision Avoidance) \cite{orca} pedestrian model for training.
%During training, at each episode, the robot is placed randomly in one of the four scenarios, and it will go through all four scenarios before repeating the set of four scenarios again. 
%All algorithms were trained using the CrowdSim\cite{rgl} simulator environment. 
% Robot max speed is set to 1.5 times max normal human walking speed.
The network parameters are taken from the original papers, and training parameters from \cite{rgl}. Both training and evaluation scenarios had five pedestrians in total. The models initially go through the same imitation learning phase using ORCA to speed up initial training. 
%Details are documented in Table \ref{tab:shared_parameters}. 
Each model is trained for 10000 episodes. The training curves were manually inspected by plotting the reward over the training episodes, and all four models were confirmed to have converged. The reward function was adopted directly from the selected papers, and all models share the same reward function.

\begin{table}[b]
    \centering
    \vspace{3mm}
    \caption{\label{tab:original}comparison of an example algorithm showing evaluation results of the original algorithm vs results after retraining using our protocol.}
    \begin{tabular}{|p{20mm}|cccc|}
    \hline
        CADRL
        & Success \%
        & Collision \%
        & Timeout \%
        & Nav. time (s) \\
    \hline
    Original ~\cite{cadrl}  
        & 86.10
        & 8.35
        & 5.55
        & 6.82 \\
    \hline
    Retrained (ours)
        & 93.5
        & 5.05
        & 1.45
        & 6.65 \\
    \hline
    \end{tabular}
\end{table}

\begin{table*}[]
\vspace{3mm}
    \centering    \caption{\label{tab:results_metrics}Test results of chosen algorithms with respect to our metrics. Tested for 500 cases for each of the 4 training scenarios in the invisible-robot environment. Bold fonts indicate best performance. Results are only taken from successful test cases. $\epsilon$ set to 0.2m to match person radius used in the tested algorithms. Only $M_{III}$ is evaluated in the visible-robot environment}
    \begin{tabular}{|c| ccccccccc|}
        \hline
        & \multicolumn{9}{c|}{Evaluation Metric} \\ \cline{2-10}
        \multirow{-2}{*}{Algorithm}
        & Success \%
        & Collision \%
        & Timeout \%
        & Nav. time (s)
        & \textbf{$M_{I}$}
        & \textbf{$M_{II}$}
        & \textbf{$M_{III} (s)$}
        & \textbf{$M_{IV} (m/s^3)^2$} 
        & \textbf{$M_{V}$} \\
        \hline
%        CADRL (default)~\cite{cadrl}  
%                            & 86.10
%                            & 8.35
%                            & 5.55
%                            & 6.82
%                            & 0.36 $\pm$ 0.35
%                            & 3.42 $\pm$ 1.61 
%                            & 141.98 $\pm$ 35.71
%                            & 8.09 $\pm$ 5.48
%                            & 3.83 $\pm$ 0.26 \\
        CADRL~\cite{cadrl}  
                    & 93.5
                    & 5.05
                    & \textbf{1.45}
                    & 6.65
                    & 0.40 $\pm$ 0.37
                    & 3.54 $\pm$ 1.50 
                    & 154.39 $\pm$ 32.23
                    & \textbf{5.63} $\pm$ 3.098
                    & 3.78 $\pm$ 0.34 \\
        GA3C-CADRL~\cite{cadrlv2}  
                            & \textbf{95.25}
                            & \textbf{2.10}
                            & 2.65
                            & 7.63
                            & \textbf{0.15} $\pm$ 0.22 
                            & \textbf{2.45} $\pm$ 1.02 
                            & 151.86 $\pm$ 39.38
                            & 10.88 $\pm$ 6.77
                            & \textbf{3.62} $\pm$ 0.33 \\
        OM-SARL~\cite{sarl}       
                            & 93.30
                            & 4.65
                            & 2.05
                            & 7.15
                            & 0.30 $\pm$ 0.32 
                            & 3.00 $\pm$ 1.33 
                            & 142.95 $\pm$ 33.00 
                            & 9.77 $\pm$ 5.59
                            & 3.67 $\pm$ 0.31 \\
        RGL~\cite{rgl}        
                            & 94.30
                            & 3.85
                            & 1.85
                            & \textbf{6.61}
                            & 0.53 $\pm$ 0.49 
                            & 2.99 $\pm$ 1.26
                            & \textbf{138.40} $\pm$ 27.51 
                            & 9.22 $\pm$ 5.32
                            & 3.86 $\pm$ 0.21 \\
        \hline
    \end{tabular}
    \vspace{-0.3cm}
\end{table*}

\begin{table}    
\caption{Test results of selected algorithms for the metric $M_{VI}$ ``Side Preference". A higher preference to one side is favored.}\label{tab:results_side}
    \begin{tabular}{|c|cc|cc|cc|}
    \hline
    \multirow{2}{*}{Algorithm} & 
    \multicolumn{2}{c|}{Passing} & 
    \multicolumn{2}{c|}{Overtaking} & 
    \multicolumn{2}{c|}{Crossing} \\ 
    \cline{2-7} & Left\% & Right\% & Left\% & Right\% & Left\% & Right\% \\ \hline
    CADRL \cite{cadrl}
        & 5 & 95 & 62 & 38 & 40 & 60 \\
    GA3C-CADRL \cite{cadrlv2}                 
        & 3 & 97 & 36 & 64 & 94 & 6 \\
    RGL \cite{rgl}                    
        & 39 & 61 & 70 & 30 & 15 & 85 \\ \hline
    \end{tabular}
\end{table}

\section{RESULT AND DISCUSSION}
        \label{sec:results}

Table \ref{tab:original} shows a comparison of the evaluation results using our protocol for an example algorithm as published (i.e., trained only in the Random scenario) \cite{cadrl}, and after retraining using our protocol. Test results after retraining for all algorithms are documented in Tables \ref{tab:results_metrics} and \ref{tab:results_side}. Table \ref{tab:original} and \ref{tab:results_metrics} present mean and standard deviation from successful test cases out of 2000 cases. %, 500 cases for each of the first four scenarios. 
Table \ref{tab:results_side} similarly presents results from successful cases for the three single-human scenarios out of 600 cases. Success rates are not reported in a separate table as the only failed cases from Table \ref{tab:results_side} are from GA3C-CADRL in the overtaking scenarios (84\% success). OM-SARL is not included in Table \ref{tab:results_side}, as the trained model does not support single-human environments.

% All algorithms are trained in an environment in which the robot's action space has a limit, and this limit is the same as human's speed limit. Thus, demonstration of the ``Walking Speed" metric will not be included in the results section. However, this metric is still an important part of our proposed metrics and will be left for future work. \\

Table \ref{tab:original} reveals that, since the original algorithm was trained and evaluated using a limited number of scenarios, it perform more poorly when encountering new scenarios. This affirms the need for and importance of a set of common evaluation protocol. Although some may think that it is "unfair" to evaluate an algorithm in a scenario that is different from the one that it was trained in, our finding brings attention to a commonly neglected but important point when training crowd navigation algorithms. All these crowd navigation algorithms aim to enable robots to navigate in not just one specific scenario, but more general in the presence of humans. As such, it is important to consider a wide variety of navigation scenarios during evaluation. Considering this, our proposed evaluation protocol covers a broad range of common navigation scenarios. Results in Table \ref{tab:original} demonstrates that using our proposed protocol for training can improve performance across broader range of scenarios.

It is interesting to note that the original CADRL presented in \cite{cadrl} was trained in the random scenario, and the random scenario should theoretically cover all scenarios in the proposed protocol. However, surprisingly, the resulting performance can still deteriorate when evaluated explicitly in our proposed scenarios. The cause for this might be that more orderly scenarios (parallel and perpendicular traffic) rarely occurs in the random generated scenario. Hence, training only in random scenario does not allow the algorithm to learn about more orderly scenarios. However, such scenarios are common in real life (at pedestrian crossing, sidewalks), and hence, important to account for.

Focusing on results after retraining as presented in Table \ref{tab:results_metrics}, we observe that all algorithms were trained successfully on the scenarios of interest, as the success rates achieved are similar to those reported in the original papers \cite{sarl, rgl}. CADRL has the lowest success rate due to its limited capability in handling multiple humans. Among the selected algorithms, RGL achieves lowest average navigation time, supporting results reported in \cite{rgl}. Our results also confirm that RGL achieves the best efficiency. CADRL achieves second-best navigation time, yet this result is overshadowed by a lower success rate.
    
Further analyzing results from our metrics, we see that although each algorithm excels at a specific aspect of social navigation, none achieves full social conformity as defined by the proposed set of metrics. For instance, GA3C-CADRL performs best with respect to the ``Personal Space" and ``Projected Path" criteria. However, among all algorithms, it has the worst ``Aggregated Time". This might be explained by GA3C-CADRL's ability to anticipate human trajectories during planning. Thus, while the resulting robot motions respect the personal space of surrounding humans, they are also overly cautious and require longer navigation times.

Similarly, our results show that RGL produces the most aggressive navigation behavior, which is why it achieves lowest navigation time. RGL also surpasses all other algorithms and achieves the lowest ``Aggregated Time". This might be due to the incorporation of a relational graph model that allows the robot to reason about its interactions with other humans and/or between humans during planning. In terms of naturalness, CADRL achieves the smoothest trajectories among all algorithms at the cost of the highest number of social zones' violations and the lowest success rate. 

Finally, with regards to the sociability of the robot's trajectories, the results in Table \ref{tab:results_side} indicate that only the algorithms, i.e., CADRL and GA3C-CADRL, that explicitly include a reward function that targets left-hand or right-hand social norms achieved a consistent side preference. The remaining two algorithms show poor preference consistency when subjected to small human position deviations.

Our results demonstrates the utility of our proposed set of metrics in providing a more comprehensive evaluation of crowd navigation methods, and better identifying their strengths and weaknesses. In terms of practical application, based on the results, if a more considerate robot was desired, such as for elder care applications, one may opt to choose GA3C-CADRL, but if a more efficient (but more aggressive) robot was desired (and acceptable), perhaps for package delivery applications, one may opt to choose RGL instead.

\noindent\textbf{Limitations:} We have evaluated the algorithms using simulation, like many existing works have done\cite{rgl,cadrlv2,sacadrl}, such that we can evaluate on a much large number of trials. In general to crowd navigation evaluation, it would be important to also evaluate in a real world setting, as there are sim-to-real gaps (especially between simulated and real human behaviors). However, this is more of a challenge to crowd navigation in general, and independent to our proposed metrics and protocol, as they can equally be used for evaluation in real world settings. In this paper, we have assumed that the robot is traveling by itself. Hence, we do not expect to have a human(s) that might be travelling closely along side the robot. If we are considering target applications where the robot is expected to travel closely beside a human(s) (e.g., a companion robot), the scenarios in the proposed protocol and the metrics would need to be modified to account for this. For instance, $M_{I}$ should not penalize having the companion human(s) close to the robot. 

% \section{LIMITATIONS}
%         \label{sec:limitations}
%         \input{Text/limitations}
        
\section{CONCLUSION}
        \label{sec:conclusion}
        In this paper, we have proposed a set of metrics and evaluation protocol that aims to facilitate more comprehensive evaluation of crowd navigation algorithms with respect to social conformity. The reason for the choice behind each metric is explained, and a demonstrated use of the metrics on several state-of-the-art algorithms is provided. We have also demonstrated that existing algorithm may fail to generalize to different scenarios, due to highly limited training/evaluating scenarios used, and that using our proposed protocol for training can improve performance across broader range of scenarios. The evaluation results highlight the ability of our presented metrics to further differentiate algorithms that have a clear ranking with respect to their efficiency, and can serve as a guideline for more socially oriented implementations and comprehensive evaluations of future algorithms. While we have proposed a set of metrics that accounts for both robot performance and social conformity, this is not an exhaustive list. Additional metrics can potentially be used in conjunction depending on the application of interest.

\bibliographystyle{IEEEtran}
\bibliography{ref}

\end{document}